\documentclass[lettersize,journal]{IEEEtran}
\usepackage{amsmath,amsfonts}
\usepackage{algorithmic}
\usepackage{algorithm}
\usepackage{array}
\usepackage[caption=false,font=normalsize,labelfont=sf,textfont=sf]{subfig}
\usepackage{textcomp}
\usepackage{stfloats}
\usepackage{url}
\usepackage{verbatim}
\usepackage{graphicx}
\usepackage{cite}
\usepackage{amsmath,amsfonts}
\usepackage{algorithmic}
\usepackage{xcolor}

\usepackage{amssymb}
\usepackage{adjustbox}
\usepackage{booktabs}
\usepackage{multirow}
\usepackage{tablefootnote}
\usepackage{hyperref}
\usepackage{float}

\usepackage{fontawesome} 
\usepackage{marvosym}

\begin{document}

\title{Training-Free Dual Hyperbolic Adapters\\ for Better Cross-Modal Reasoning}


\author{Yi~Zhang,~Chun-Wun~Cheng,~Junyi~He,~Ke~Yu,~Yushun~Tang,~Carola-Bibiane Schönlieb, Zhihai He, and~Angelica~I.~Aviles-Rivero\textsuperscript{\Letter}
\thanks{Yi Zhang is with the College of Computer Science and Software Engineering, Shenzhen University, Shenzhen, China (e-mail: rambo.ai@szu.edu.cn).}
\thanks{Junyi He, Ke Yu, Yushun Tang and Zhihai He are with the Department of Electrical and Electronic Engineering, Southern University of Science and Technology, Shenzhen, China (e-mail: \{hejy2021, yuk2020, tangys2022\}@mail.sustech.edu.cn; hezh@sustech.edu.cn).}
\thanks{Chun-Wun Cheng and Carola-Bibiane Schönlieb are with the Department of Applied Mathematics and Theoretical Physics, University of Cambridge, Cambridge, UK (e-mail: \{cwc56, cbs31\}@cam.ac.uk).}
\thanks{ Angelica I. Aviles-Rivero is with the Yau Mathematical Sciences Center, Tsinghua University, Beijing, China (e-mail: aviles-rivero@tsinghua.edu.cn).}
\thanks{\Letter Corresponding author: Angelica I. Aviles-Rivero.}
}

\markboth{}%
{Zhang \MakeLowercase{\textit{et al.}}: Training-Free Dual Hyperbolic Adapters for Better Cross-Modal Reasoning}


\maketitle

\begin{abstract}
Recent research in Vision-Language Models (VLMs) has significantly advanced our capabilities in cross-modal reasoning. However, existing methods suffer from performance degradation with domain changes or require substantial computational resources for fine-tuning in new domains. To address this issue, we develop a new adaptation method for large vision-language models, called \textit{Training-free Dual Hyperbolic Adapters} (T-DHA). We characterize vision-language relationship between semantic concepts, which typically has a hierarchical tree structure, in the hyperbolic space instead of the traditional Euclidean space. Hyperbolic spaces exhibit exponential volume growth with radius, unlike the polynomial growth in Euclidean space. We find that this unique property is particularly effective for embedding hierarchical data structures using the Poincaré ball model, achieving significantly improved representation and discrimination power. Coupled with negative learning, it provides more accurate and robust classifications with fewer feature dimensions. Our extensive experimental results on various datasets demonstrate that the T-DHA method significantly outperforms existing state-of-the-art methods in few-shot image recognition and domain generalization tasks. 

\end{abstract}


\section{Introduction}
\label{sec:intro}
\IEEEPARstart{L}{arge} Vision-Language Models (VLMs), such as CLIP~\cite{radford2021learning} and ALIGN~\cite{jia2021scaling}, are trained on extensive image-text datasets using contrastive learning. These models excel in creating a unified vision-language embedding space by aligning visual and textual modalities, enabling their successful application across a wide range of downstream visual tasks, such as few-shot image recognition~\cite{zhang2022tip,zhou2022learning,zhu2023not}. However, when VLMs are applied to new domains, domain shifts can widen the gap between the vision and language modalities, necessitating adaptation to obtain satisfactory performance. To enhance the transferability of VLMs, researchers have proposed efficient adaptation techniques, including prompt tuning methods such as CoOp~\cite{zhou2022learning}, CoCoOp~\cite{zhou2022conditional}, and ProDA~\cite{lu2022prompt}, as well as adapter-style approaches like CLIP-Adapter~\cite{gao2021clip} and GraphAdapter~\cite{li2024graphadapter}.

\begin{figure}[!t]
    \centering
    \includegraphics[width = 1\linewidth]{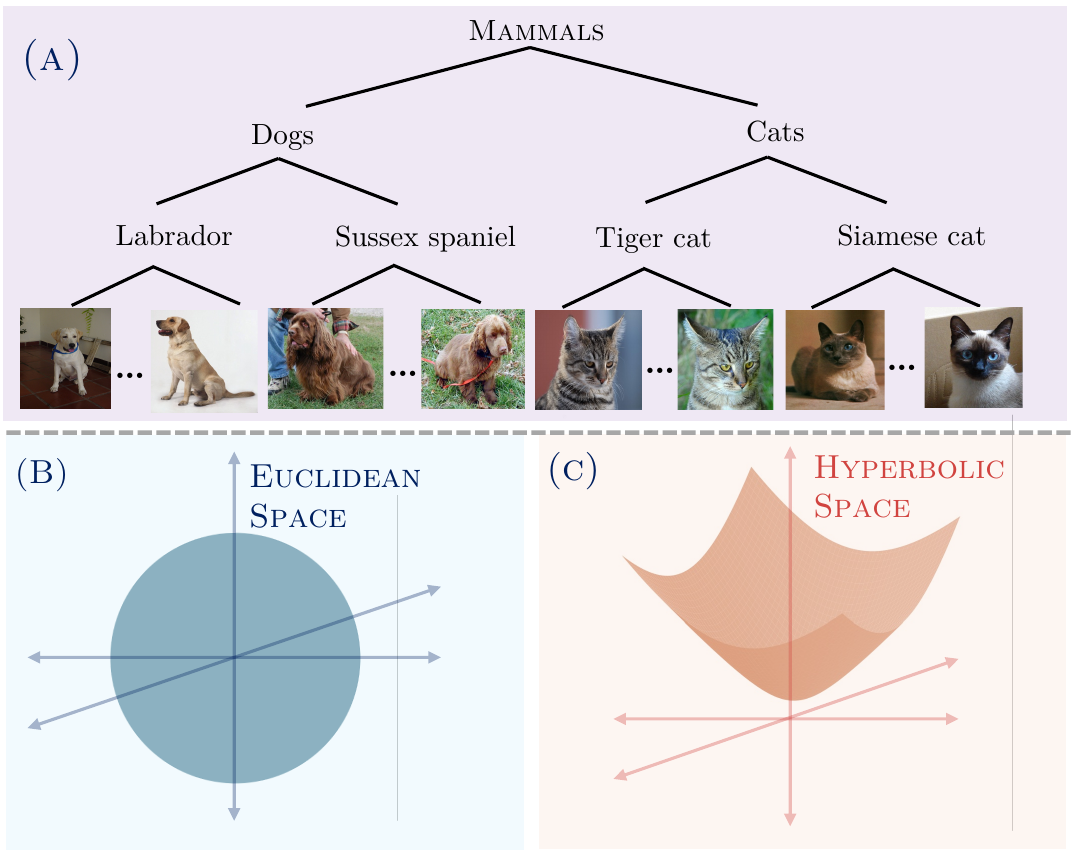}
    \caption{\textbf{Hierarchical Data Representation in Euclidean vs. Hyperbolic Spaces.} (A) The example displays a hierarchical categorization of mammals into dogs and cats, further divided by breeds. (B) Euclidean space struggles to effectively capture hierarchical structures. (C) Hyperbolic space naturally accommodates tree-like structures, maintaining clear distinctions between hierarchical levels. This motivates our use of hyperbolic class prototypes within the T-DHA architecture (Section III-B), where embeddings are explicitly mapped into the Poincaré ball model for improved representation.}
    \label{fig:feature_map}
\end{figure}

However, compared to zero-shot CLIP~\cite{radford2021learning}, these adaptation methods require significantly more computational resources to fine-tune the newly introduced learnable parameters. For instance, CoOp takes more than 10 hours to train. As a result, training-free adaptation approaches have garnered considerable attention. Tip-Adapter~\cite{zhang2022tip}, a pioneering training-free adaptation method, leverages CLIP's zero-shot classification capabilities while achieving performance comparable to training-required methods. It employs a strategy that combines Image-Image similarity and Image-Text similarity to produce the final prediction; the Image-Text similarity is the same as in zero-shot CLIP.

In this paper, to address the problem of training-free cross-modal few-shot learning, we propose a novel method called Training-free Dual Hyperbolic Adapters (T-DHA). Inspired by Tip-Adapter, we exploit the paradigm that fuses Image-Image prediction and Image-Text prediction to compute the final prediction. However, we introduce two major improvements: 1) employing hyperbolic distance for measuring image similarity, and 2) introducing negative learning for both image-image prediction and image-text prediction. 

\textit{\textbf{Why do we use hyperbolic distance as the similarity metrics? }} 
While methods such as Tip-Adapter rely on cosine similarity in Euclidean space, we propose the use of hyperbolic geometry for training-free few-shot learning. Our approach is motivated by several key advantages of hyperbolic spaces in modeling semantic relationships.

First, hyperbolic space exhibits exponential volume growth with radius, in contrast to the polynomial growth of Euclidean space. This property makes it particularly suitable for embedding hierarchical structures, thereby enhancing representational capacity. As demonstrated by Sarkar et al.\cite{sarkar2011low}, tree-like data can be embedded with low distortion in a Poincaré disk, which naturally preserves hierarchical relations. For example, in a hierarchical classification task such as “Mammal” → “Dog” → “Labrador”, hyperbolic embeddings more accurately capture the semantic relationships, as illustrated in Figure~\ref{fig:feature_map}. By keeping “Labrador” closer to “Dog” and farther from “Cat”, hyperbolic geometry improves generalization from limited examples and supports more robust hierarchical inference.

\begin{figure*}[ht]
\begin{center}
\centerline{\includegraphics[width=0.8\linewidth]{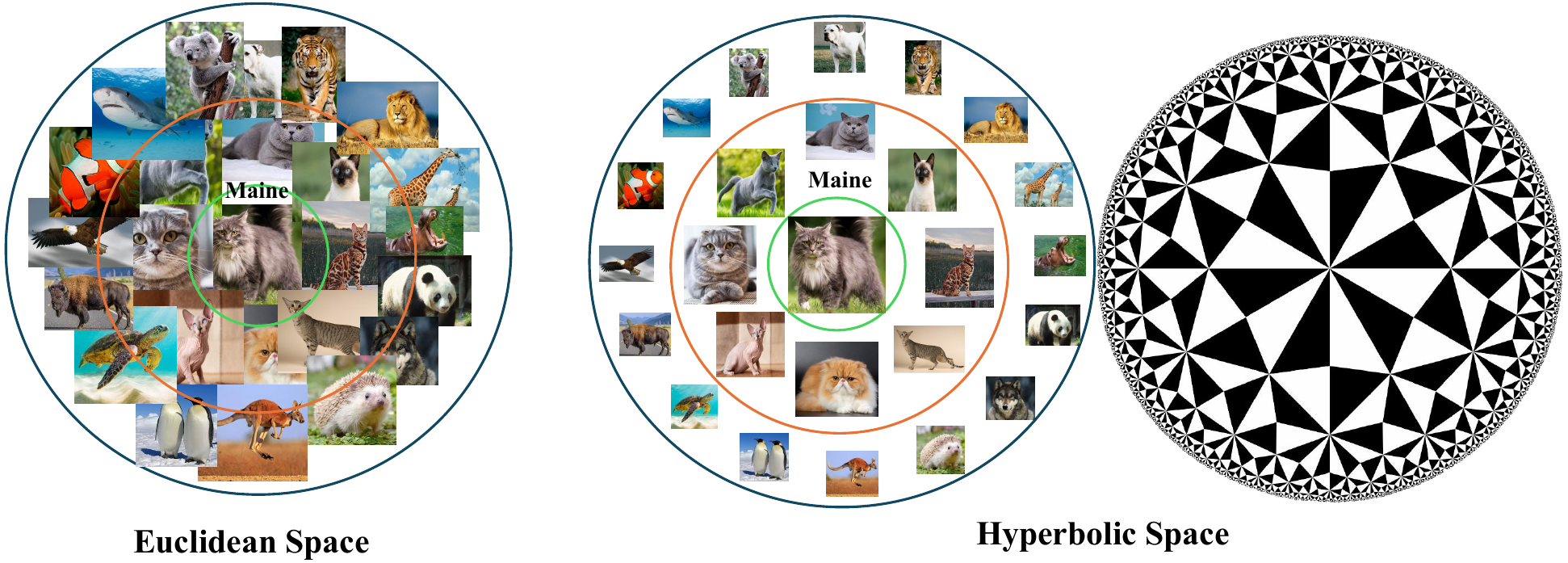}}
\caption{\textbf{Embedding Images in a Conceptual Hierarchy.} 
 Euclidean geometry (a) struggles to model the exponential growth of semantic concepts around a central node (e.g., a Maine Coon cat). In contrast, hyperbolic geometry (b) provides a native structure for such hierarchies, with volume expanding exponentially to incorporate all concepts. The right-hand side figure in hyperbolic geometry (b) has been taken from MathWorld, Wolfram Research ~\cite{mathworld_poincare_disk}}

\label{fig:euc_vs_hyperbolic}
\end{center}
\end{figure*}

Second, hyperbolic space offers a geometrically principled inductive bias for semantic hierarchies inherent in visual recognition tasks. This advantage can be broken down into a clear logical chain:
1) Inherent semantic hierarchy: class labels (e.g., Animal → Mammal → Dog → Labrador) naturally form a tree-structured taxonomy. 2) CLIP encodes hierarchy: CLIP’s text encoder maps such hierarchical class prompts into a shared embedding space where geometric relations reflect semantic relations.
3) Image–Text alignment: through contrastive training, CLIP aligns image embeddings with corresponding text concepts. An image of a Labrador is embedded near the textual concepts “a photo of a Labrador”, “a photo of a dog”, and transitively, “a photo of a mammal”. 4) Hyperbolic space as optimal inductive bias: given the hierarchical nature of the semantic space, a Euclidean metric such as cosine similarity is suboptimal. Hyperbolic space, with its exponential growth, offers a natural geometric prior that preserves and leverages this semantic structure during training-free adaptation, as visualized in Fig.~\ref{fig:euc_vs_hyperbolic}.

Third, hyperbolic embeddings enable effective low-dimensional representation without sacrificing discriminative power. In few-shot settings where efficiency is critical, Euclidean dimensionality reduction often leads to blurred class boundaries and overlapping representations. In contrast, hyperbolic space maintains clear separation between semantically similar classes, allowing compact yet discriminative representations. This ensures that even in reduced dimensions, the model can accurately distinguish between fine-grained categories such as dog breeds, maintaining high performance with minimal parameters.

\textit{\textbf{Why do we exploit negative learning?}} 
Negative learning is crucial for refining a model’s \textit{discriminatory power} by explicitly accounting for what an image is not, in addition to what it is. Consider the task of distinguishing between cats and dogs. The model evaluates an image against both the positive prototype of a cat and the negative prototype of a dog. This dual comparison is key when the image exhibits features common to both categories, such as pointy ears and small noses. By recognizing these similarities, the model can judiciously adjust its confidence levels. For instance, if an image aligns closely with the cat prototype yet shares certain canine traits, the model's confidence in its initial cat classification is moderated, thereby reducing potential misclassifications. This method ensures that decision-making is not solely based on feature matching but includes a deliberate process to exclude incorrect classifications. Thus, negative learning is crucial in enhancing classification accuracy and robustness, especially in scenarios where different classes share overlapping features. Without this approach, models may fail to recognize subtle yet critical differences, leading to errors in classification.

Based on the reasoning outlined above, we utilize hyperbolic distance to measure similarity in image-image prediction and apply negative learning to enhance performance in both image-image and image-text predictions. Specifically, in positive image-image prediction, we calculate the hyperbolic distance between the test image and the positive hyperbolic class prototypes. For negative image-image prediction, we compute the hyperbolic distance between the test image and the negative hyperbolic class prototypes, where the negative prototype of a certain class is derived from images of other classes within the same dataset.
For image-text prediction, we treat CLIP's zero-shot prediction as the positive prediction, achieved by calculating the cosine similarity between the image features and the text embeddings generated from prompts like ``a photo of \{class\}". The negative image-text prediction is similar, except we introduce negation by adding ``no" or ``not" before the class name, altering the prompt to ``a photo of no \{class\}". The final prediction is a combination of these four predictions.
Our approach shows that both hyperbolic distance and negative learning contribute significantly to the improved performance of our method. We conducted extensive experiments to evaluate the proposed T-DHA on few-shot image recognition and domain generalization tasks. The comprehensive empirical results demonstrate that T-DHA significantly outperforms existing state-of-the-art methods.

\section{Related Work}
\label{sec:related work}

\begin{figure*}[ht]
\begin{center}
\centerline{\includegraphics[width=1\linewidth]{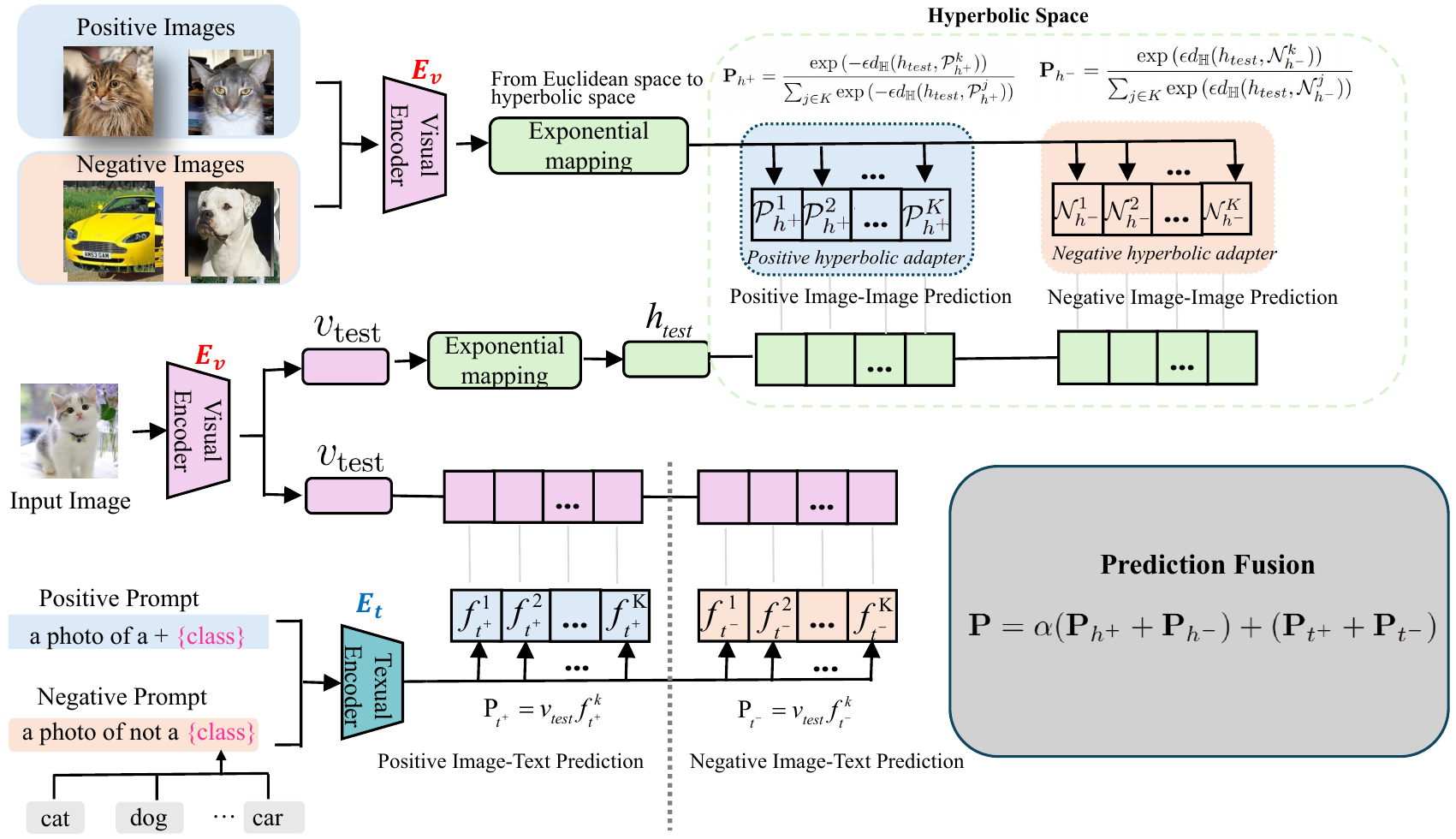}}
\caption{\textbf{Overview of the Proposed Training-free Dual Hyperbolic Adapters (T-DHA)}. 
The architecture leverage image-image and image-text predictions. For image-image prediction, visual features are first extracted using the CLIP encoder and then mapped into hyperbolic space via the exponential map (Equation 2). Predictions are computed using Poincaré distance to both positive and negative class prototypes in hyperbolic space. The image-text branch similarly integrates positive and negative predictions via prompt-based cosine similarity. These geometry-aware branches are fused to yield the final prediction. All geometric operations are now annotated in the figure and fully described in Section III-B.}
\label{fig:overview}
\end{center}
\end{figure*}

\noindent\textbf{Adaptation of Vision-Language Models.}
Vision-language models (VLMs) have proven highly effective in learning semantic relationships between images and text through extensive pre-training on large-scale datasets~\cite{radford2021learning, jia2021scaling, lu2019vilbert, yu2022coca, li2021supervision}. Notable examples like CLIP~\cite{radford2021learning} and ALIGN~\cite{jia2021scaling} have leveraged contrastive learning to align visual and textual embeddings, enabling robust zero-shot learning and other open-world tasks. To fine-tune these pre-trained VLMs for specific downstream tasks, two primary approaches have emerged: prompt tuning and feature adaptation. Specifically, prompt tuning methods focus on optimizing textual or visual prompts to better align with the target task while preserving the core VLM parameters~\cite{zhou2022learning, zhou2022conditional, chen2022prompt, ma2023understanding, shu2022tpt, lu2022prompt, zhang2024unleash, liu2024generalizable}. Feature adaptation methods, on the other hand, add lightweight adapters to directly fine-tune the representations produced by VLMs~\cite{gao2021clip, zhang2022tip, yu2023task, li2024graphadapter, udandarao2023sus, zhang2024cross, he2024vlab}. Beyond these, other innovative strategies for VLM adaptation, such as using LLMs to generate customized prompts~\cite{pratt2023does, menon2022visual} and cross attention~\cite{guo2023calip, weng2025cross}, have also been explored.

\vspace{5pt}
\noindent\textbf{Hyperbolic Space.} 
Hyperbolic spaces have gained prominence for effectively modeling hierarchical and tree-like structures, which are challenging to capture using Euclidean spaces. Foundational works such as Poincaré embeddings~\cite{nickel2017poincare} and subsequent advancements in hyperbolic neural networks (HNNs)\cite{ganea2018hyperbolic, gulcehre2018hyperbolic, chami2019hyperbolic, liu2019hyperbolic} demonstrated the suitability of hyperbolic geometry for representing complex hierarchical data. These principles have been extended to various applications, including natural language processing~\cite{zhu2020hypertext, chen2021probing}, image processing~\cite{ahmad2022fisheyehdk, zhang2022hyperbolic, cui2023rethinking}, and more. In the realm of VLMs, the recent work MERU~\cite{desai2023hyperbolic} has harnessed hyperbolic embeddings to improve cross-modal tasks, enabling more efficient and interpretable representations. HyCoCLIP ~\cite{pal2024compositional} learns a shared hyperbolic embedding space by combining contrastive and entailment objectives to hierarchically align whole images, object-level boxes, and their corresponding text descriptions. HySAC ~\cite{pal2024compositional} fine-tunes CLIP in hyperbolic space with contrastive and entailment-cone losses to anchor safe image-text pairs near the origin and push unsafe pairs outward, enabling safety-aware retrieval traversals and an implicit NSFW classifier. Unlike MERU, which trains models in hyperbolic space, our approach uses hyperbolic space directly for similarity computation, enhancing hierarchical representation while maintaining efficiency.

\section{Method}
\label{sec:method}
\subsection{Preliminaries}
\noindent\textbf{(1) CLIP~\cite{radford2021learning}}
consists of an image encoder ($E_v$), typically using architectures like ResNet \cite{he2016deep} or Vision Transformer (ViT) \cite{dosovitskiy2020image}, and a text encoder ($E_t$) that employs a Transformer architecture.
In an image classification task, we are given a test image $X_{test}$ belonging to a class $y$, where $X_{test}\in \mathbb{R}^{C\times H\times W}$ and $y \in \mathbb{R}^K$ for a classification problem with $K$ classes. In zero-shot CLIP, each class label $y_i$ in the class set $Y=\{y_1, y_2, \cdots, y_K\}$ is combined with a predefined prompt, such as $\rho =$ ``a photo of", to form textual inputs for the various classes, expressed as $\{\rho; y_i\}$. The text encoder $E_t$ then produces text embeddings $\{t_1, t_2, \cdots, t_K\}$, where each $t_i = E_t({\rho; y_i})$. Next, a cosine similarity score is calculated by pairing each text embedding $t_i$ with the image feature $v=E_v(X_{test})$, enabling the determination of the likelihood that $X_{test}$ is associated with class $y_i$: 
$\mathrm{sim}\left( t_i,v \right)=\frac{t_i \cdot v}{\Vert t_i \Vert  \Vert v \Vert}$.
The prediction probability on $X_{test}$ can be denoted by 
\begin{equation}
\label{eq-clip}
   p(y_i|X_{test})=\frac{\exp \left( \mathrm{sim}\left( t_i,v \right) /\tau \right)}{\sum\nolimits_{j=1}^K{\exp \left(\mathrm{sim}\left( t_j,v \right) /\tau \right)}}, 
\end{equation}
where $\tau$ refers to the temperature of the softmax function.

\vspace{5pt}
\noindent\textbf{(2) Hyperbolic Space.} 
Hyperbolic space, integral to hyperbolic geometry and a subset of Riemannian geometry, is crucial for understanding our T-DHA. This space involves manifolds—generalizations of curved surfaces into higher dimensions and more complex structures—central to differential geometry. Riemannian geometry reveals that manifolds display hyperbolic geometry when each point has a consistently negative curvature, giving it a characteristic saddle-like shape at every point.
At any point \( a \) on a manifold \( \mathcal{M} \), the tangent space \( T_a\mathcal{M} \) consists of all possible directions for tangential movement at \( a \). Within this space, the inner product is defined explicitly. The Riemannian metric \( k \), which is essentially a set of these inner products, is defined as \( k_x : T_x\mathcal{M} \times T_a\mathcal{M} \rightarrow \mathbb{R} \), for \( a \in \mathcal{M} \). Once \( k \) is established, the pair \( (\mathcal{M}, k) \) is defined as a Riemannian manifold.
It is important to first introduce the following definitions: 1) \textbf{Geodesic} is the shortest path between two points on a curved surface,
2) \textbf{Parallel Transport} is the process of moving a vector along a curve on a surface or manifold while keeping it parallel to itself according to the connection or curvature of that space. Mathematically, it is defined as $P_{a \rightarrow b} : T_a\mathcal{M} \rightarrow T_{b}\mathcal{M}$ 3) \textbf{Exponential Map} maps from $T_a\mathcal{M}$ to $\mathcal{M}$ and provides a way to project from Euclidean space to hyperbolic space. Consider any tangent vector $w \in T_0\mathcal{M} \setminus \{0\}$. The exponential map $\exp_{0}$ reads:  
\begin{equation}
\label{eq: exp}
    \exp_0(w) = \tanh(\|w\|) \frac{w}{\|w\|}.
\end{equation}
We select $\mathbf{0}$ as our reference point. 
Note that an exponential map lies in its ability to project image features between two distinct spaces: the Euclidean space and the hyperbolic space. 
The inverse operation of the exponential map is the logarithmic map $\log_0(y) : \mathcal{M} \rightarrow T_0\mathcal{M}$. Clearly, $\log_0(\exp_0(v)) = v$ holds true.

\vspace{5pt}
\noindent\textbf{(3) Poincaré Ball.} 
In our work we use the Poincaré ball model to represent visual features such that all the points are inside a unit ball. A Riemannian manifold can be represented as \((\mathbb{H}^n, k^{\mathbb{H}})\), where \(\mathbb{H}^n = \{ a \in \mathbb{R}^n : \|a\| < 1 \}\) denotes the \(n\)-dimensional hyperbolic space within the Poincaré ball model. The Riemannian metric on the Poincaré ball, denoted by \(k^{\mathbb{H}}\), is defined at each point \( a \) by the expression: $k^{\mathbb{H}}_a = \lambda_a^2 k^E$, where the conformal factor \(\lambda_a\) is given by $\lambda_a = (1 - \|a\|^2)^{-1}$. Here, \(k^E\) represents the Euclidean metric. This conformal factor \(\lambda_a\) adjusts the Euclidean metric to account for the curvature of the hyperbolic space, providing a consistent way to measure distances and angles within the Poincaré ball.
The Poincaré distance between two points \((a, b)\) can be derived using the Riemannian metric as follows:
\begin{equation}
    \label{eq:h_dict}
    d_{\mathbb{H}}(a, b) = \cosh^{-1} \left( 1 + 2 \frac{\|a - b\|^2}{(1 - \|a\|^2)(1 - \|b\|^2)} \right)
\end{equation}

\noindent\textbf{(4) Relationship between Hyperbolic space and the Poincaré ball}
The Poincaré ball model is one of the most widely used formulations of hyperbolic space because it provides a conformal mapping — meaning that angles are preserved — which is crucial for maintaining local geometric structure when computing similarities. This property allows distances to grow exponentially toward the boundary of the ball, naturally reflecting the hierarchical expansion present in many real-world label structures.

\subsection{Training-free Dual Hyperbolic Adapters (T-DHA)}
\noindent\textbf{(1) Overview} As depicted in the Figure~\ref{fig:overview}, we present our Training-free Dual Hyperbolic Adapters (T-DHA). T-DHA integrates both image-image and image-text predictions. In image-image prediction, positive and negative hyperbolic adapters are utilized, containing class prototypes respectively. For a given image \( x_{\text{test}} \), its feature \( v_{\text{test}} \) is extracted using the encoder \( E_v \) and then mapped into hyperbolic space as \( h_{\text{test}} \). Hyperbolic distances to both positive and negative class prototypes are computed to generate respective predictions, which are then combined to form the final image-image output.
In image-text prediction, we enhance CLIP's zero-shot technique by utilizing discriminative text features derived from class-specific prompts, adjusted with "not" or "no" for negative predictions. These positive and negative predictions are combined to form the image-text prediction.

\vspace{5pt}
\noindent\textbf{(2) Positive Hyperbolic Adapter for Image-Image Prediction.}
The Positive Hyperbolic Adapter is a parameter-free linear layer model that contains the hyperbolic class prototypes of all the classes within a dataset. Next, we explain how to obtain the hyperbolic class prototypes and perform the image-image prediction.
Given the pre-trained CLIP $\{E_v;E_t\}$ and a new dataset with $N$-shot $K$ classes, each class contains $N$ annotated images. The training set is represented as $X_{train} \triangleq \{x_j\}_{j=1}^{NK}$, where $N \in \{1, 2, 4, 8, 16\}$. By using the CLIP visual encoder $E_v$, we generate the image features $V_{train} \triangleq \{v_j\}_{j=1}^{NK}$, where $v_j = E_v(x_j)$. For each class $k$ in $K$, the initial class prototype is obtained by averaging the $N$ image features, denoted as,
\begin{equation}
    \mathcal{P}_{k}=\frac{1}{N}\sum_{n=1}^N{v}_n.
\end{equation}
To align with the geometry of hierarchical semantic concepts, we project the Euclidean class prototypes into hyperbolic space using the exponential map. This transformation is central to our architecture and is now also visually indicated in Figure 2. Once in the Poincaré ball, similarity computations are conducted via hyperbolic distance (Equation 3), rather than Euclidean or cosine similarity.
Next, we use Equation~\ref{eq: exp} to map the initial prototype to the Poincaré ball, namely from Euclidean space to hyperbolic space, using the exponential map, denoted as,
\begin{equation}
\label{eq: exp_prototype}
    \mathcal{P}_{h^{+}}^k = \tanh(\|\mathcal{P}_k\|) \frac{\mathcal{P}_k}{\|\mathcal{P}_k\|},
\end{equation}
In this way, we obtain all the positive hyperbolic class prototypes $\hat{\mathcal{P}}_{h^{+}} \triangleq \{\mathcal{P}_{h^{+}}^k\}_{k=1}^K$. Given a test image $x_{test}$, we can measure the similarity between $x_{test}$ and the positive hyperbolic class prototypes $\hat{\mathcal{P}}_{h^{+}}$ using Equation~\ref{eq:h_dict}. Therefore, for the positive image-image prediction, given a test image $x_{test}$, we first extract the image feature $v$ using $E_v$, and then utilize Equation~\ref{eq: exp_prototype}
to map $v$ into hyperbolic space, resulting in $h_{test}$. 
The probability of $x_{test}$ belonging to class $k$ is:
\begin{align}
\label{eq:h_poitive}
\mathbf{P}_{h^{+}}(y=k|x_{test}) &= \frac{\exp{(-\epsilon d_{\mathbb{H}}(h_{test}, \mathcal{P}_{h^{+}}^k))} }
    {\sum_{j\in K} \exp{(-\epsilon d_{\mathbb{H}}(h_{test}, \mathcal{P}_{h^{+}}^j))} }.    
\end{align}
Here, following \cite{zhang2024training}, we introduce a hyperparameter $\epsilon$, denoted as a temperature factor.

\vspace{5pt}
\noindent\textbf{(3) Negative Hyperbolic Adapter for Image-Image Prediction}
We next explain why we use Negative Hyperbolic Prototypes and introduce the details of the negative hyperbolic adapter for image-image prediction.

\textit{\textbf{What are the key differences between the negative learning approach proposed in T-DHA and the negative handling mechanism in CLIP?}} our proposed negative learning in T-DHA is fundamentally different from the implicit negative handling in contrastive pre-training, in both timing and formulation:
\textbf{1. Stage of Application: }  CLIP leverages negatives only during large-scale pre-training, where the negatives are sampled from the batch and serve to shape the global embedding space. In contrast, our negative learning is applied at inference time within a training-free adaptation framework. This means we explicitly incorporate negative prototypes at test time without additional training or fine-tuning, which is not the case for CLIP or typical contrastive methods.
\textbf{2. Explicit Prototype-Based Negatives: } Rather than relying on incidental negatives in a contrastive loss, we construct explicit negative prototypes for both image-image and image-text predictions. These prototypes are computed from other classes in the same target dataset (for image-image) or from negated textual prompts such as “a photo of no {class}” (for image-text). This formulation allows the model to directly measure “how much an input is not a class,” rather than relying solely on relative similarity among batch elements.
\textbf{3. Dual Positive-Negative Fusion: } Our method fuses four signals:
1). Positive image-image similarity (hyperbolic distance to positive prototypes). 
2). Negative image-image similarity (hyperbolic distance to negative prototypes). 
3). Positive image-text similarity (cosine similarity to standard prompts). 
4). Negative image-text similarity (cosine similarity to negated prompts)
This explicit dual-positive-negative combination enables the model to adjust confidence scores more precisely in challenging few-shot and domain-shift scenarios, especially when inter-class features overlap.

\textit{\textbf{Why do we need negative hyperbolic prototypes?}}
Incorporating negative hyperbolic prototypes into classification models enhances their discriminatory ability by assessing non-membership, reducing false positives, and improving accuracy. For instance, a cat image is compared not only to its class prototype but also to negative prototypes from the dog and bird classes. This dual assessment system ensures more accurate classifications by confirming positive identifications and rejecting negative ones, especially beneficial in few-shot learning scenarios.
Consider a test image hyperbolic feature \(h_{\text{cat}}\) representing a cat, and class prototypes: positive prototypes \(\mathcal{P}_{h^{+}}^{\text{cat}}, \mathcal{P}_{h^{+}}^{\text{dog}}, \mathcal{P}_{h^{+}}^{\text{bird}}\) and negative hyperbolic prototypes \(\mathcal{N}_{h^{-}}^{\text{cat}}, \mathcal{N}_{h^{-}}^{\text{dog}}, \mathcal{N}_{h^{-}}^{\text{bird}}\). The logit score for the cat class is computed by combining positive similarity (high similarity to the cat prototype) and dissimilarity to the negative prototype (low similarity to non-cat prototypes). If \(h_{\text{cat}}\) is close to \(\mathcal{P}_{h^{+}}^{\text{cat}}\) and far from \(\mathcal{N}_{h^{-}}^{\text{cat}}\), the combined logit score will be high for the cat class, leading to confident classification. This illustrates how negative hyperbolic prototypes refine the classification process by clearly identifying non-membership, thus enhancing model performance.

Next, we introduce the details of the negative hyperbolic adapter. For class $k \in K$ of a dataset, we first construct its negative hyperbolic prototype $\mathcal{N}_{h^{-}}^k$ calculated as the average hyperbolic feature of images that do not belong to class $k$. In detail, we randomly select a single image from the other $K-1$ classes and compute the average of their hyperbolic features to represent the negative hyperbolic prototype of class $k$~\cite{zhang2024negative}. This process yields the negative prototypes for the target dataset, denoted as $\hat{\mathcal{N}}_{h^{-}} \triangleq \{\mathcal{N}_{h^{-}}^k\}_{k=1}^K$. For example, in a dataset with classes {cat, dog, bird}, to create the negative hyperbolic prototype for the cat class, we randomly choose one single image from the dog and bird classes. We then extract their image feature ($v_{dog}$, $v_{bird}$)  using $E_v$ and map these features into hyperbolic space using
Equation~\ref{eq: exp_prototype} to obtain their hyperbolic features ($h_{dog}$, $h_{bird}$).
The negative hyperbolic prototype of cat is then calculated as
$\mathcal{N}_{h^{-}}^{\text{cat}} = \frac{h_{dog} + h_{bird}}{2}$. Therefore, for the negative hyperbolic adapter, the probability of $x_{test}$ belonging to class k can be computed by
\begin{align}
\label{eq:h_negative}
\mathbf{P}_{h^{-}}(y=k|x_{test}) &= \frac{\exp{(\epsilon d_{\mathbb{H}}(h_{test}, \mathcal{N}_{h^{-}}^k))} }
    {\sum_{j\in K} \exp{(\epsilon d_{\mathbb{H}}(h_{test}, \mathcal{N}_{h^{-}}^j))} }.    
\end{align}
The construction of negative hyperbolic prototypes in our method uses a Euclidean average of selected negative class features, followed by an exponential map into the Poincaré ball. We emphasise that this is a deliberate and commonly used approximation in hyperbolic few-shot learning  \cite{ganea2018hyperbolic, chami2019hyperbolic, zhang2022hyperbolic}. While it does not yield a true Riemannian Fréchet mean, this choice ensures computational efficiency and simplicity — particularly important in our training-free, inference-only setup. For small support sizes, the local geometry near the origin of the Poincaré ball approximates Euclidean behavior, making this projection sufficiently accurate. Empirical results (Section IV) confirm that this approximation yields robust performance However, all layers in \cite{van2023poincare}: convolutions, residual blocks, ReLU, batchnorm — every operation performed on the Poincaré ball. In our method, Post-feature stage only: CLIP embeddings are mapped from Euclidean to hyperbolic space via exponential map for similarity computation. Then we will provde an explanation on why the Euclidean mean followed by exponential projection closely approximates the Fr´echet mean. The Riemannian mean $\mu$ of a set of points $\{x_i\}_{i=1}^N$ 
minimizes the total squared geodesic distance
\[
\mu = \arg\min_{z \in \mathcal{M}} \sum_{i=1}^N d(z, x_i)^2,
\]
and satisfies the well--known first--order stationarity condition
\[
\sum_{i=1}^N \log_\mu(x_i) = 0.
\]
Linearizing around $\mu \simeq 0$ (or, more generally, in a small neighborhood
of a base point) and using the local approximation 
$\log_0(x_i) \approx x_i$, we obtain
\[
\frac{1}{N}\sum_{i=1}^N \log_0(x_i) \;\approx\;
\frac{1}{N}\sum_{i=1}^N x_i.
\]
Hence, a first--order estimator of the mean in the tangent space is
\[
\hat{\mu} = \exp_0\!\left(\frac{1}{N}\sum_{i=1}^N \log_0(x_i)\right)
\;\approx\;
\exp_0\!\left(\frac{1}{N}\sum_{i=1}^N x_i\right).
\]
This provides a first order approximation to the true Fréchet mean.

Therefore, we can obtain overall image-image prediction by,


\begin{equation}
\label{eq:iip}
    \mathbf{P}_{II}(y=k|x_{test}) = \mathbf{P}_{h^{+}}(y=k|x_{test}) + \mathbf{P}_{h^{-}}(y=k|x_{test})
\end{equation}

\vspace{5pt}
\noindent\textbf{(4) Positive Image-Text Prediction}
The positive image-text prediction is analogous to the zero-shot CLIP reference. We first construct a prompts bank $\Omega \triangleq \{\omega_k\}_{k=1}^K$ for all $K$ classes in a dataset, each $\omega$ contains $L$ prompts. 
These prompts are collected from CLIP~\cite{radford2021learning} and CuPL~\cite{pratt2023does}. The prompts from CLIP are hand-crafted templates such as ``a photo of \{class\}", while the prompts from CuPL are customized, higher accuracy prompts tailored to the task domain and generated by LLMs. For example, for the platypus class, the prompts might be ``A platypus looks like a beaver with a duck's bill", ``The platypus is an odd-looking animal with a duck-like bill, beaver-like tail, and otter-like body." To mitigate the prompts bias, we compute the average of the textual features of the prompts for each class to perform the prediction, denoted as $f_{t^{+}}^k = \frac{1}{L}\sum_{l=1}^L{E_t(PMT_l)}$, where PMP denotes the prompt and $L$ is the number of prompts for class $k$. Using this, for the positive image-text prediction, the prediction of the test image $x_{test}$ belonging to class $k$ is given by:
\begin{equation}
\mathbf{P}_{t^{+}}(y=k|x_{test}) = \frac{ \exp(\text{sim}(v, f_{t^{+}}^k)/\tau)}{\sum_{j=1}^{K}\ \exp(\text{sim}(v, f_{t^{+}}^j))/ \tau))},
\label{eq:positive_prob}
\end{equation}
where $v$ is the image feature of $x_{test}$ generated by $E_v$.

\vspace{5pt}
\noindent\textbf{(5) Negative Image-Text Prediction.} Motivated by the same reason we use the negative hyperbolic adapter, we also leverage the negative image-text prediction. The process is similar to positive image-text prediction, but the difference lies in the construction of prompts. We add ``no" or ``not" in the prompts like ``a photo of no \{class\}" to form the negative prompts. Therefore, we can get the negative image-text prediction of the text image $x_{test}$ belonging to class $k$, denoted as:
\begin{equation}
\mathbf{P}_{t^{-}}(y=k|x_{test}) = \frac{ \exp(-\text{sim}(v, f_{t^{-}}^k)/\tau)}{\sum_{j=1}^{K}\ \exp(-\text{sim}(v, f_{t^{-}}^j))/ \tau))},
\label{eq:negative_prob}
\end{equation}
where $f_{t^{-}}^k$ represents the negative textual feature of class $k$. The overall image-text prediction can be obtained by:
\begin{equation}
\label{eq:itp}
    \mathbf{P}_{IT}(y=k|x_{test}) = \mathbf{P}_{t^{+}}(y=k|x_{test}) + \mathbf{P}_{t^{-}}(y=k|x_{test})
\end{equation}

\vspace{5pt}
\noindent\textbf{(6) Prediction Fusion.} Finally, according to Equation~\ref{eq:iip} and Equation~\ref{eq:itp}, the ultimate predicted probability of the input image $x_{test}$ is given by:
\begin{equation}
    \label{eq-LOGIT}
    \begin{split}
        \mathbf{P}(y=k|x_{test}) = &\ \alpha\mathbf{P}_{II}(y=k|x_{test}) + \mathbf{P}_{IT}(y=k|x_{test})
    \end{split}
\end{equation}
where $\alpha$ is used to control the scaling of the residual connection. $\mathbf{P}(y=k|x_{test})$ is the final prediction.

\section{Experiments}

\subsection{Experimental Settings}
We adopted the experimental configurations used in Tip-Adapter~\cite{zhang2022tip} for evaluating few-shot learning and domain generalization. Our experiments were conducted on 11 visual recognition datasets: Caltech101~\cite{fei2004learning}, DTD~\cite{cimpoi2014describing}, EuroSAT~\cite{helber2019eurosat}, FGVCAircraft~\cite{maji2013fine}, Flowers102~\cite{nilsback2008automated}, Food101~\cite{bossard2014food}, ImageNet~\cite{deng2009imagenet}, OxfordPets~\cite{parkhi2012cats}, StanfordCars~\cite{krause20133d}, SUN397~\cite{xiao2010sun}, and UCF101~\cite{soomro2012ucf101}. These datasets cover a range of categories, including generic objects, actions, scenes, fine-grained categories, textures, and remote sensing, providing a comprehensive evaluation of our approach. We used the few-shot evaluation protocol from Tip-Adapter~\cite{zhang2022tip} and APE~\cite{zhu2023not}, randomly selecting 1, 2, 4, 8, and 16 shots from training sets and evaluating on the test sets. For domain generalization, we evaluated our method's robustness to domain shifts using ImageNetV2~\cite{recht2019imagenet} and ImageNet-Sketch~\cite{wang2019learning}, which test its ability to generalize across different domains.

\vspace{5pt}
\noindent\textbf{Baselines.} We compare our proposed approach against the following state-of-the-art approaches: zero-shot CLIP~\cite{radford2021learning},Tip-Adapter~\cite{zhang2022tip}, APE~\cite{zhu2023not}, CALIP~\cite{guo2023calip}, Tip-X~\cite{udandarao2023sus}, and DMN-TF~\cite{zhang2024dual}, ECALP~\cite{li2025efficient}. 
To ensure a fair comparison, the baselines results were directly taken from their respective research papers.


\subsection{Implementation Details}
Our method is developed using the publicly accessible CLIP model as its foundation. Building on previous research~\cite{zhou2022conditional,zhang2022tip}, we utilize a ResNet-50~\cite{he2016deep} backbone as our visual encoder and employ a transformer as the text encoder in our default experimental setup. We adopt prompt ensembling, leveraging textual prompts from both CLIP~\cite{radford2021learning} and CuPL~\cite{pratt2023does} to enhance model performance and adhere to the data pre-processing protocol outlined in CLIP for all datasets. To ensure the reliability of our results, we repeat each experiment three times with varying initialization seeds and report the average accuracy. All experiments are performed on a single NVIDIA RTX 3090 GPU.

\subsection{Performance Analysis}
\label{subsec:percom}

\begin{figure*}[h]
\centering
\includegraphics[width=0.95\textwidth]{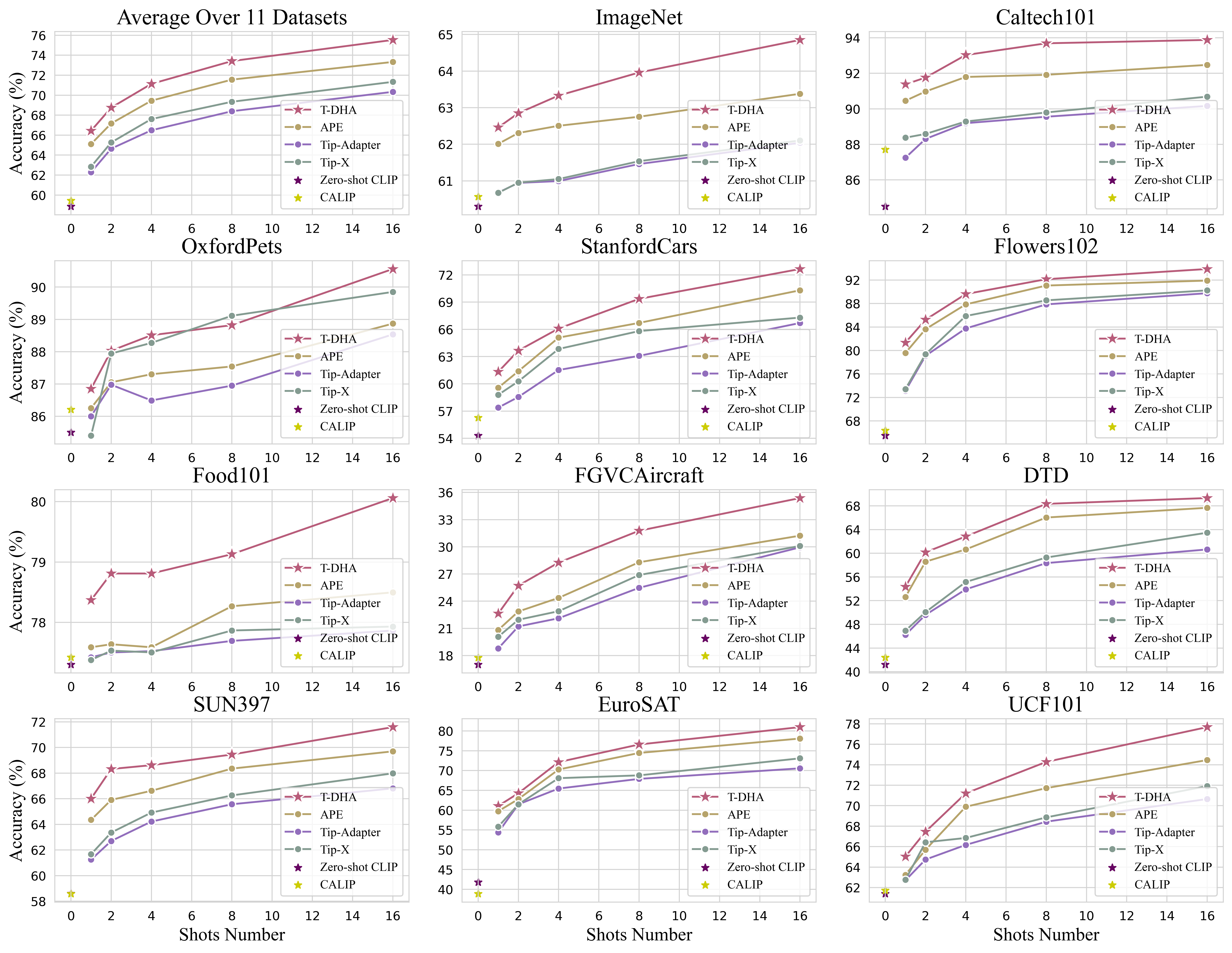}
\caption{\textbf{Classification Performance Comparison on Training-free Few-shot Learning}, {\em i.e.}, 1-/2-/4-/8-/16-shot, on 11 benchmark datasets. The top-left is the averaged accuracy across all 11 datasets.}
\label{fig:free_fewshot_results_dha}
\end{figure*}

\vspace{5pt}

\begin{table} [t!]
\centering
\caption{\textbf{Classification Performance Comparison on Training-free Few-shot Learning. We report the average of 11 datasets}}
\label{table:add_refs}
\begin{adjustbox}{width=\linewidth}
\begin{tabular}{l|cccccc}
\toprule
Method   & 1-shot   & 2-shot & 4-shot  & 8-shot & 16-shot \\
\midrule
Tip-Aapter~[\textit{ECCV2022}] & 62.29   & 64.65  & 66.49 & 68.39   & 70.33 \\
APE~[\textit{ICCV2023}]   & 65.11   & 67.17  & 69.45 & 71.55   & 73.33 \\
DMN-TF~[\textit{CVPR2024}]   & 66.26  & 67.77 & 69.86 & 71.28   & 73.15 \\
ECALP~[\textit{ICLR2025}] & 65.83   & 68.62 & 70.46 & 71.64   & 74.03 \\
\textbf{T-DHA (Ours)} & \textbf{66.44}   & \textbf{70.31} & \textbf{71.19} & \textbf{73.41}   & \textbf{75.53} \\
\bottomrule
\end{tabular}
\end{adjustbox}
\end{table}

\noindent\textbf{(1) Training-Free Few-Shot Recognition.}  
We summarize the experimental results in Figure~\ref{fig:free_fewshot_results_dha} and Table~\ref{table:add_refs}. In the top-left plot, we present the average accuracy over 11 datasets. It is obvious that our proposed T-DHA significantly and consistently outperforms other state-of-the-art methods. For example, compared to APE~\cite{zhu2023not}, T-DHA achieves a performance gain of 1.33\%, 1.60\%, 1.8\%, 2.0\%, and 2.3\% at 1, 2, 4, 8, and 16 shots, respectively. Our method also achieves substantial improvements over other baselines, such as a 4.2\% increase over Tip-X~\cite{udandarao2023sus} in the 16-shot setting. For the largest dataset in our experiments, ImageNet, our T-DHA surpasses APE by 1.5\%. Notably, for challenging tasks such as FGVC Aircraft and UCF101, our method gains a performance boost of 4.16\% and 3.22\% compared to APE, respectively. Table~\ref{table:add_refs} compares our method with existing approaches in terms of average accuracy across 11 datasets. Our proposed T-DHA consistently and significantly outperforms the second-best method (ECALP~\cite{li2025efficient}), achieving a 1.5\% performance improvement in the 16-shot setting.
These performance comparisons provide experimental evidence demonstrating the strong few-shot adaptation capability in a training-free manner, highlighting the effectiveness of the hyperbolic embeddings and positive-negative learning strategy for few-shot recognition.

\vspace{5pt}
\noindent\textbf{(2) Domain Generalization.} 
The domain generalization task is utilized to evaluate our model's capability to generalize to a target domain that differs from the source domain. As shown in Table~\ref{imagenett}, we compare our method with four state-of-the-art methods, including zero-shot approaches \cite{radford2021learning,guo2023calip} and training-free methods \cite{zhang2022tip,zhu2023not}. Our approach consistently outperforms all the compared models by a significant margin across two out-of-distribution datasets. Compared to the second-best method, APE \cite{zhu2023not}, T-DHA achieves up to a 1.17\% improvement on ImageNet-V2 and a 1.31\% improvement on ImageNet-Sketch. When compared to Tip-Adapter, the performance gains increase to 2.51\% on ImageNet-V2 and 2.02\% on ImageNet-Sketch. These results demonstrate that our T-DHA is remarkably robust to distribution shifts.

\begin{table}[t!]
\centering
\caption{Performance Comparisons on Domain Generalization}
\begin{adjustbox}{width=\linewidth}
\begin{tabular}{lcccc}
\toprule
\multirow{3}{*}{Methods} & \multirow{3}{*}{Training Type} & \textbf{Source} &\multicolumn{2}{c}{\textbf{Target}} \\
\cmidrule(lr){3-3} \cmidrule(lr){4-5} & 
& ImageNet  & -V2 & -Sketch \\ \midrule
CLIP  & \multirow{2}*{Zero-shot} &60.33  & 53.27 & 35.44\\
CALIP  &  & 60.57  & 53.70 & 35.61\\ 
\midrule
Tip-Adapter & \multirow{3}*{Training-free}  &{62.03}  &  {54.60} &  {35.90} \\
APE& & 63.42  & 55.94 & 36.61 \\
T-DHA (Ours) &  &  \textbf{64.85} &\textbf{57.11} & \textbf{37.92} \\
\bottomrule
\end{tabular}
\end{adjustbox}
\label{imagenett}
\end{table}

\subsection{Ablation Studies}

In this section, we perform comprehensive ablation studies to evaluate the impact of various components of our method. Unless stated otherwise, all experiments are conducted on the 16-shot ImageNet dataset.

\vspace{5pt}
\noindent\textbf{(1) Effectiveness of Different Components.} 
Our method consists of four components: Positive/Negative hyperbolic image-image prediction ($IIP_p$ and $IIP_n$) and Positive/Negative image-text prediction ($ITP_p$ and $ITP_n$). Positive image-text prediction aligns with the common zero-shot CLIP approach. The last row in Table~\ref{table:comp} shows our full T-DHA architecture, highlighting how each additional component enhances performance. Notably, the model achieves a 3.47\% improvement when comparing the second row to the last, underscoring the effectiveness of hyperbolic representations. Additionally, comparisons between the first and second rows, and the third and last rows, show that negative prediction significantly boosts performance. These results confirm the value of our hyperbolic embeddings and negative prediction strategy. Incorporating negative learning in this explicit, prototype-driven manner yields significant accuracy improvements beyond what is achievable with CLIP’s original zero-shot predictions or with positive-only prototype matching. This demonstrates that our negative learning is not redundant with CLIP’s contrastive pre-training, but rather provides complementary discriminative information at inference time.

\begin{table}[t!]
\caption{Effectiveness of Different Algorithm Components in T-DHA\tablefootnote{$ITP_p$ and $ITP_n$ refer to Image-Text Positive and Negative Prediction, respectively, while $IIP_p$ and $IIP_n$ represent Image-Image Positive and Negative Prediction, respectively.}}
\label{table:comp}
\centering
\resizebox{\linewidth}{!}{
\begin{tabular}{lccccc}
\toprule
\multirow{2}{*}{Method} & \multicolumn{5}{c}{Number of Shots}          \\
\cmidrule(lr){2-6}
                        & 1     & 2     & 4     & 8     & 16    \\
\midrule
 $ITP_p$ (Zero-shot CLIP)  & 60.33 & 60.33 & 60.33 & 60.33 & 60.33 \\
$ITP_p$ + $ITP_n$                & 60.53 & 60.88 & 61.09 & 61.12 & 61.38\\
$ITP_p$ + $ITP_n$ +$IIP_P$               & 61.85 & 62.43 & 62.85 & 63.48 & 64.02 \\
$ITP_p$ + $ITP_n$ +$IIP_p$ + $IIP_n$ (Ours)   & \textbf{62.01} & \textbf{62.85} & \textbf{63.33} & \textbf{63.96} & \textbf{64.85} \\    
\bottomrule
\end{tabular}
}
\end{table}

\vspace{5pt}
\noindent\textbf{(2) Evaluation on Various Visual Backbones.}
We conduct experiments on various backbones, including ResNets~\cite{he2016deep} and ViTs~\cite{dosovitskiy2020image}, using the 16-shot setting on ImageNet. The results are presented in Table~\ref{table:backbones}. Remarkably, our proposed T-DHA achieves significant performance improvements over other methods across all visual backbones, demonstrating its overall effectiveness and versatility.

\begin{table}[t!]
\caption{Evaluation Across Various Visual Backbones}
\label{table:backbones}
\centering
\begin{adjustbox}{width=\linewidth}
\begin{tabular}{lccccc}
\toprule
\multirow{2}{*}{Method} & \multicolumn{5}{c}{Visual Backbone} \\
\cmidrule(lr){2-6} & ResNet-50 & ResNet-101 & ViT-B/32 & ViT-B/16 & ViT-L/14 \\
\midrule
Zero-shot CLIP  & 60.33 & 62.53 & 63.80  & 67.83  & 75.43  \\ 
Tip-Adapter           & 62.03     &  64.78      & 65.61   & 70.75  & 76.19   \\
T-DHA (Ours) & \textbf{64.85}     & \textbf{66.03}      & \textbf{67.97}    & \textbf{72.64}   &\textbf{79.16} \\
\bottomrule
\end{tabular}
\end{adjustbox}

\end{table}

\vspace{5pt}
\noindent\textbf{(3) Residual Ratio $\alpha$.} 
The hyperparameter $\alpha$ determines the balance between the hyperbolic image-image prediction and the image-text prediction of the pre-trained CLIP model. It can also be seen as the weight for hyperbolic image-image similarity in Equation~\ref{eq:iip}. A larger $\alpha$ indicates greater reliance on hyperbolic image-image similarity. As shown in Table~\ref{tb:hyper}, classification accuracy improves as $\alpha$ increases from 0.0 to 1.2, peaking at 64.85\% when $\alpha = 1.2$. This suggests that hyperbolic image-image similarity has a more significant impact on the final prediction than CLIP's text-image similarity.
In the Supplemental Materials, we include details about our proposed method and additional experimental results.

\vspace{5pt}
\noindent\textbf{(4) Euclidean Cosine Similarity VS Hyperbolic Distance.} 
we conduct an ablation where the hyperbolic distance in our method was replaced with Euclidean cosine similarity. The results (Table~\ref{table:ECS_vs_HD}) show consistent accuracy drops  under identical settings. This supports our claim that hyperbolic distance offers a measurable advantage in our training-free few-shot setting.
\begin{table} [h!]
\centering
\caption{\textbf{Classification Performance Comparison on Euclidean Cosine Similarity(ECS) and Hyperbolic Distance(HD). We report the average over 11 datasets.}}
\label{table:ECS_vs_HD}
\begin{adjustbox}{width=\linewidth}
\begin{tabular}{l|cccccc}
\toprule
Method   & 1-shot   & 2-shot & 4-shot  & 8-shot & 16-shot \\
\midrule
ECS & 65.27   & 69.02 & 69.58 & 71.65  & 73.76 \\
\textbf{HD (T-DHA)} & \textbf{66.44}   & \textbf{70.31} & \textbf{71.19} & \textbf{73.41}   & \textbf{75.53} \\
\bottomrule
\end{tabular}
\end{adjustbox}
\end{table}

\vspace{5pt}
\noindent\textbf{(5) Positive and Negative Pipeline Analysis. }To evaluate their impact, we conducted an ablation study by removing each pipeline individually. As shown in Table~\ref{table:Pos_vs_Neg}, removing the negative pipeline results in accuracy drops of 0.82\% (1-shot), 1.56\% (2-shot), 1.02\% (4-shot), 1.98\% (8-shot), and 1.52\% (16-shot), demonstrating its importance in refining decision boundaries. Removing the positive pipeline causes larger drops, confirming its central role in correct classification. Using both pipelines yields the highest accuracy, indicating that they work synergistically to enhance discriminative power.

This quantitative analysis clearly demonstrates the necessity and effectiveness of the proposed dual-pipeline design. 
\begin{table} [h!]
\centering
\caption{\textbf{Classification Performance Comparison on Positive(Pos) and Negative(Neg) pipeline. We report the average over 11 datasets.}}
\label{table:Pos_vs_Neg}
\begin{adjustbox}{width=\linewidth}
\begin{tabular}{l|cccccc}
\toprule
Method   & 1-shot   & 2-shot & 4-shot  & 8-shot & 16-shot \\
\midrule
Neg & 62.17   & 64.82 & 66.18 & 69.05  & 70.16 \\
Pos & 65.62   & 68.75 & 70.17 & 71.43  & 74.01 \\
\textbf{Pos + Neg (T-DHA)} & \textbf{66.44}   & \textbf{70.31} & \textbf{71.19} & \textbf{73.41}   & \textbf{75.53} \\
\bottomrule
\end{tabular}
\end{adjustbox}
\end{table}

\begin{figure*}[ht]
\begin{center}
\centerline{\includegraphics[width=0.8\linewidth]{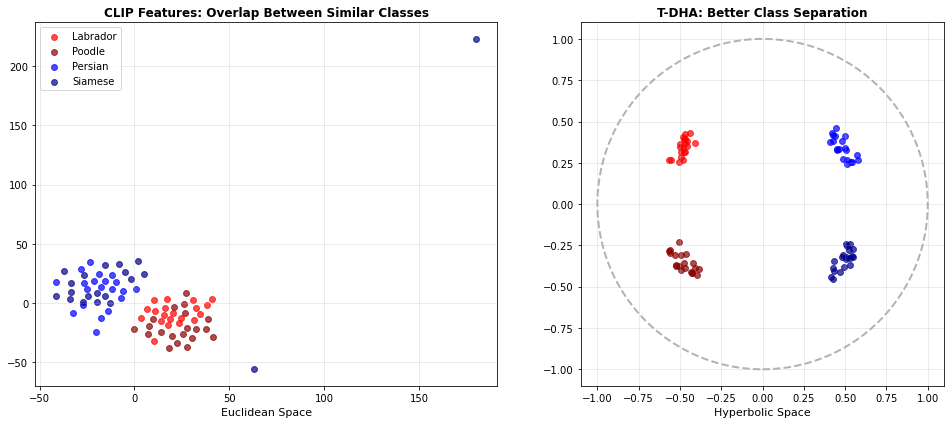}}
\caption{\textbf{Visualization comparison of feature distributions in Euclidean vs. hyperbolic spaces.}.  Left: CLIP features in Euclidean space (t-SNE projection) show significant overlap between similar classes, making classification challenging. Right: T-DHA features in hyperbolic space (Poincaré disk) achieve clear separation of all classes, demonstrating the advantage of hyperbolic geometry for hierarchical data. Colors represent different classes: Red/Light Red = Dog breeds (Labrador/Poodle), Blue/Light Blue = Cat breeds (Persian/Siamese). The hyperbolic representation better preserves semantic hierarchies, enabling more accurate few-shot classification.}
\label{fig:clip_vs_hyperbolic}
\end{center}
\end{figure*}

\noindent\textbf{(6) Visualization comparison of feature distributions in Euclidean vs. hyperbolic spaces.} 
As shown in the figure~\ref{fig:clip_vs_hyperbolic}:
Left panel (Euclidean Space): Features from CLIP (processed through t-SNE) demonstrate the inherent challenges in few-shot learning. Similar classes (e.g., Labrador vs. Poodle, Persian vs. Siamese) exhibit significant overlap, making accurate classification difficult. This overlap reflects the limitations of Euclidean geometry in modeling hierarchical relationships.
Right panel (Hyperbolic Space): Our T-DHA method produces embeddings that better preserve hierarchical structures. The Poincaré disk visualization shows clear separation between all classes, with each category maintaining distinct clusters. This demonstrates hyperbolic space's superior capability in capturing semantic relationships.

\vspace{5pt}
\noindent\textbf{(7) Efficiency Comparisons.} 
We conduct timing experiments to quantitatively compare our approach with existing methods(Tip-Adapter, APE, DMN-TF, ECALP). We measure the wall-clock time on 16-shot ImageNet. The results are shown in the Table~\ref{table:efficiency}. We can see that our proposed T-DHA chieves the best performance with comparable computational cost.

\begin{table}[h!]
\caption{Sensitivity of Hyperparameters}
\centering
\begin{adjustbox}{width=0.95\linewidth}
	\begin{tabular}{c|cccccc}
	\toprule
		\multicolumn{7}{c}{Sensitivity of Hyperparameters} \\ 
		\midrule
		$\alpha$  &0.0 &0.4&0.8 &\textbf{1.2} &1.6 &2.0 \\  
        \cmidrule(lr){1-7}
		 Acc. &61.38  & 63.18  &64.46 &\textbf{64.85} &\text{64.31} &63.23\\ 
	    
	\bottomrule
	\end{tabular}
\end{adjustbox}
\label{tb:hyper}
\end{table}

\begin{table} [h!]
\centering
\caption{\textbf{Efficiency comparisons on 16-shot ImageNet}. We report the results using a single NVIDIA RTX A6000 GPU.}
\label{table:efficiency}
\begin{adjustbox}{width=\linewidth}
\begin{tabular}{l|cccccc}
\toprule
Method   & Train    & Test & GFLOPs  & Param. & Acc. \\
\midrule
Tip-Aapter~[\textit{ECCV2022}] & -   & 10.4ms  & - & -   & 62.03 \\
APE~[\textit{ICCV2023}]   & -   & 10.4ms  & - & -   & 63.38 \\
DMN-TF~[\textit{CVPR2024}]   & -  & 10.7ms & - & -   & 64.28 \\
ECALP~[\textit{ICLR2025}] & -   & 10.6ms & - & -   & 64.37 \\
\textbf{T-DHA (Ours)} & -   & \textbf{10.2ms} & - & -   & \textbf{64.85} \\
\bottomrule
\end{tabular}
\end{adjustbox}
\end{table}

\label{sec:experiments}

\section{Conclusion and Future Work}
\label{sec:conclusion}
We introduced Training-free Dual Hyperbolic Adapters (T-DHA), a method that uses hyperbolic spaces for efficient domain adaptation in vision-language models without additional training. T-DHA, employing hyperbolic distance for image comparisons and incorporating negative learning, enhances cross-modal reasoning and outperforms existing methods in few-shot image recognition and domain generalization. Our experiments demonstrate that T-DHA surpasses current state-of-the-art methods, offering a promising solution for efficient and robust cross-modal reasoning.

 While our current work focuses on CLIP-like vision-language models for image recognition, the proposed framework is general and can, in principle, be applied to Multimodal Large Language Models (MLLMs). MLLMs typically produce joint embedding spaces for multiple modalities (e.g., image, text, audio), and our hyperbolic distance–based similarity computation can be applied to these embeddings in the same way as for CLIP features.

Specifically, the mapping from Euclidean to hyperbolic space does not depend on the encoder architecture or modality type; it only requires fixed feature vectors from the model. Therefore, any MLLM output features — whether visual, textual, or cross-modal — can be projected into the Poincaré ball and benefit from the representational advantages of hyperbolic geometry. Similarly, the positive/negative pipeline can be extended to multi-modal prompts or prototypes in MLLMs, enabling training-free adaptation in multi-modal settings.
\section*{Acknowledgments}
CWC and AIAR acknowledge support from the Swiss National Science Foundation (SNSF) under grant number 20HW-1 220785. CBS acknowledges support from the Philip Leverhulme Prize, the Royal Society Wolfson Fellowship, the EPSRC advanced career fellowship EP/V029428/1, EPSRC grants EP/S026045/1 and EP/T003553/1, EP/N014588/1, EP/T017961/1, the Wellcome Innovator Awards 215733/Z/19/Z and 221633/Z/20/Z, CCMI and the Alan Turing Institute. 
AIAR gratefully acknowledges the support of the Yau Mathematical Sciences Center, Tsinghua University. This work is also supported by the Tsinghua University Dushi Program.

\bibliographystyle{IEEEtran}
\bibliography{IEEEabrv,ref}

\end{document}